# How good is GPT at writing political speeches for the White House?

Jacques Savoy
Computer Science Dept., University of Neuchatel
Neuchâtel, Switzerland
Jacques.Savoy@unine.ch

**Abstract**

Using large language models (LLMs), computers are able to generate a written text in response to a us er request. As this pervasive technology can be applied in numerous contexts, this study analyses the written style of one LLM called GPT by comparing its generated speeches with those of the recent US presidents. To achieve this objective, the *State of the Union* (SOTU) addresses written by Reagan to Biden are contrasted to those produced by both GPT-3.5 and GPT-4.o versions. Compared to US presidents, GPT tends to overuse the lemma "we" and produce shorter messages with, on average, longer sentences. Moreover, GPT opts for an optimistic tone, opting more often for political (e.g., president, Congress), symbolic (e.g., freedom), and abstract terms (e.g., freedom). Even when imposing an author's style to GPT, the resulting speech remains distinct from addresses written by the target author. Finally, the two GPT versions present distinct characteristics, but both appear overall dissimilar to true presidential messages.

## 1   Introduction

With the development of large language models (LLMs) (Zhao *et al*., 2023), generative AI demonstrates its capability to generate a short text in response to a user request. Currently, such applications are freely available and can help users produce various types of writing (e.g., e-mail, CV, short letter, etc.). From this perspective, this study investigates the writing style of GPT when asked to generate *State of Union* addresses for a president. Annually expressed in front of Congress, these speeches explain the world situation and political agenda of the occupant of the White House. The main objective is to inform and persuade the audience that the propositions and actions of the president are the most appropriate. To reach such an objective, the style and rhetoric play an important role in reinforcing the president's words.

Based on recent developments in automated text analysis designed by communication and psychological scholars (Jordan, 2022), this study analyses the style and rhetoric of six US presidents (Reagan, Clinton, Bush, Obama, Trump, and Biden) as well as that of two GPT versions (GPT-3.5 and GPT-4.o). In this study, rhetoric is defined as the art of effective and persuasive speaking, and the way to adopt a tone to motivate an audience. An author's style is evaluated through studying frequent forms employed to support his/her communication objective (Biber & Conrad, 2009).

To author a SOTU speech, a chief ghostwriter collaborates more or less closely with the president[1]. Could we employ GPT to achieve a similar objective and expect that it could adopt a political tone and style of the current occupant of the White House? In the end, can we still discriminate between the generated address and the real one? If so, what are the stylistic characteristics that differ between the two speeches? Moreover, what are the rhetoric features that can be pertinent to discriminate between the addresses written by several presidents (Reagan, Clinton, Bush, Obama, Trump, and Biden)? Additionally, can we observe distinct aspects between the two GPT versions and, if so, which one is the best to write a political message?

To address these questions, this article is organised as follows. The first section presents some related work, while Section 3 describes the corpus used in our experiments. Section 4 analyses some stylistic features by comparing those in both GPT versions to those occurring in speeches written by US presidents. Additional experiments focusing on psychological and emotional characteristics are depicted in Section 5, while the next evaluates the global similarity between each president and the two GPT versions. Finally, a conclusion reports the main findings of this study.

## 2  State of the Art

Numerous studies have been published on authorship attribution and on recognising author demographics characteristics (e.g., gender, age, social status, native language, etc.) (Kreuz, 2023). Other stylometry studies have additionally been performed on the detection of plagiarism or fake documents, the identification of suspects in criminology (Olsson, 2018), the determination of text genre, and even the dating of a document. To resolve these questions, various natural language processing models have been applied by scientists from different domains such as computer science (Savoy, 2020), (Karsdorp *et al*., 2021), linguistics (Crystal, 2019), (Yule, 2020), psychology (Pennebaker *et al*., 2014), (Jordan, 2022) and communication studies (Hart *et al*., 2013), (Hart, 2020).

The main objective of this study is to analyse the style and rhetoric of true political speeches and to compare them with those automatically generated by GPT. This emerging technology is based on LLM (large language model) technology grounded on a deep learning architecture (Goodfellow *et al*., 2016), which is based on a sequence of transformers with an attention mechanism (Vaswami *et al.* 2017). The most important notion to understand LLM is the following: given a short sequence of tokens (e.g., words or punctuation symbols), the computer is able to automatically supply the next token. More precisely, knowing four tokens, the model must first determine the list of possible next tokens to complete the given sequence (Wolfram, 2023). For example, after the chain "the president of the," the computer, based on the training documents, can define a list of the next occurring token, such as United, Philippines, Senate, US, USA, UK, republic, Ukraine, and so forth.

---

[1] For example, Obama's ghostwriter, J. Favreau, comments his job at www.youtube.com/watch?v=zFbaesLEa4g

From this list, and depending on some parameters, the system can then select the most probable token (in our case, "United") or based on a uniform distribution, one over the top *k* ranked tokens (e.g., "Senate"), or randomly depending on their respective probabilities of occurrence in the training texts (e.g., "US"). This non-deterministic process guarantees that the same request will produce distinct messages. Common to all LLMs, GPT may include hallucinations in its answers (namely, incorrect information). In our previous example, the sequence "the president of the UK" would be replaced by "the Prime Minister of the UK"). Moreover, the specification of the sources exploited to produce the text stays unknown to the user[2].

As previously mentioned, the main target application of such LLMs is to generate a short text in the context of a dialogue. To analyse such automatically generated texts, different studies expose the effectiveness of several learning strategies capable of discriminating between answers generated by GPT-3.5 and answers written by human beings (Guo *et al.*, 2023). Based on a classifier trained on a given domain (e.g., RoBERTa), the recognition rate is rather high (around 95% to 98%). Such effectiveness is also obtained when the target language is not English (e.g., French (Antoun *et al.*, 2023)), or when it is Japanese (Mizumoto *et al.*, 2024). Such a high degree could be reduced when faced with a new and unknown domain or when substituting tokens by misspelled words (in such cases, the achieved accuracy rate varies from 28% to 60%). Of course, the message must include at least 1,000 letters to allow the detection system to reach such a small error rate.

With a similar objective, the CLEF-PAN 2019 international evaluation campaign evaluated different systems to automatically detect whether a set of tweets was generated by bots or by humans (Daelemans *et al.*, 2019). In this case as well, the effectiveness was rather high (between 93% to 95% for the best approaches). However, the tweets written by bots were not produced by a LLM, but corresponded to messages either containing a well-known citation, a passage of the Bible, or text corresponding to a predefined pattern (e.g., list of positions available in a large company).

## 3  Corpus Overview

To ground our conclusions on a solid basis, the same text genre has been selected: namely, written speeches given in the same context, to achieve similar objectives, and written in the same time period. To compare the style of recent US presidents with messages created by a machine, we asked GPT API (Application Programming Interface) to generate the *State of the Union* (SOTU) addresses for six presidents, namely Reagan, Clinton, Bush, Obama, Trump, and Biden.

---

[2] The training sample employed by GPT is not precisely known and one might assume that many presidential speeches have been included.

For each US leader, only the SOTU addresses were taken into consideration. In addition, two versions of GPT were used, namely version 3.5 and 4.o (or 4.omni). As shown in Table 1, the number of SOTU speeches varied from three (Biden) to eight (Clinton, Bush, Obama).

**Table 1.** Some statistics on our American corpus

|  | Presidency | Number | Tokens | Types | Mean length |
|---|---|---|---|---|---|
| Reagan-GPT-3.5 |  | 70 | 29,381 | 1,074 | 414.7 |
| Clinton-GPT-3.5 |  | 80 | 42,125 | 1,385 | 528.0 |
| Bush-GPT-3.5 |  | 70 | 35,756 | 1,254 | 504.9 |
| Obama-GPT-3.5 |  | 80 | 32,224 | 1,340 | 539.7 |
| Trump-GPT-3.5 |  | 40 | 19,616 | 1,033 | 484.5 |
| Biden-GPT-3.5 |  | 30 | 15,282 | 977 | 489.3 |
| Reagan-GPT-4.o |  | 70 | 45,651 | 1,221 | 643.1 |
| Clinton-GPT-4.o |  | 80 | 55,085 | 1,275 | 680.8 |
| Bush-GPT-4.o |  | 70 | 45,665 | 1,277 | 643.9 |
| Obama-GPT-4.o |  | 80 | 52,557 | 1,414 | 649.2 |
| Trump-GPT-4.o |  | 40 | 25,049 | 1,027 | 614.4 |
| Biden-GPT-4.o |  | 30 | 19,879 | 941 | 640.1 |
| R. Reagan | 1981–1989 | 7 | 32,490 | 3,384 | 3,975.4 |
| B. Clinton | 1993–2000 | 8 | 59,705 | 3,835 | 6,520.5 |
| W.G. Bush | 2001–2008 | 8 | 40,532 | 3,514 | 4,349.5 |
| B. Obama | 2009–2016 | 8 | 53,777 | 3,902 | 6,021.0 |
| D. Trump | 2017–2020 | 4 | 22,189 | 3,200 | 3,973.8 |
| J. Biden | 2021–2024 | 3 | 25,598 | 2,912 | 5,778.0 |

To help both GPT versions in their generative process[3], the true SOTU address of the corresponding year was included in the prompt. In addition, a short list of possible topics was inserted (e.g., "deregulation, free market, reduced taxes, small government, education, middle-class, security, …"). Finally, the prompt[4] specified the president's name and year to obtain a message written according to the style of a specified leader. For example, for 1982, the prompt included the following sentences:

> "I'm Ronald Reagan, President of the United States of America. I need to write my SOTU speech. Can you write a SOTU speech to be presented in the front of the Congress in January 1982"

---

[3] The training sample used by GPT is unknown but one can assume that many presidential speeches have been included. However, those messages, if appearing in the training set, are employed to define the occurrence probability of a token, given the four previous ones, and not to identify a presidential style.

[4] All the prompts are available at https://drive.switch.ch/index.php/s/rqcHexKib7QsdLS. Moreover, the parameters have been fixed as follows: temperature=0.5, frequency_penalty=0, presence_penality=0, top_p=0.4, max_tokens=32768. All the prompts are available at https://drive.switch.ch/index.php/s/rqcHexKib7QsdLS

As GPT generates relatively short messages, ten different versions for each speech have been generated for both versions. As shown in the Appendix, this limit seems problematic for OpenAI, particularly when generating political speeches.

Table 1 depicts a general overview of our political corpus. The third column indicates the number of speeches. The total number of tokens (labelled "Tokens") and the number of distinct words (labelled "Types") are reported in the next columns. These values are computed without counting the numbers and the punctuation symbols.

The last column shows the mean number of tokens per speech. The average size of the GPT versions is roughly ten times smaller than the real ones. When comparing both GPT versions, the overall mean length is 119.3 for GPT-3.5 and 658.0 with GPT-4.o, a significant difference (bilateral *t*-test, significance level 1%). In total, this corpus contains 663,871 tokens, with 429,580 created by GPT and 234,291 belonging to true SOTU addresses.

## 4   Stylometric Analysis

As a first stylometric measure, one can focus on the language complexity that all political leaders tend to reduce. For example, L. B. Johnson (presidency: 1963–1969) specifies to his ghostwriters, "I want four-letter words, and I want four sentences to the paragraph." (Sherrill, 1967). The complexity of the language could be measured by the mean number of letters per words. In this case, the larger the mean, the higher the language complexity.

As an additional characteristic, one can count the percentage of words composed of six letters or more, defined as big words (BW) in the English language. One can observe, for example, that depending of the length of words, some are easier to understand than others. It is the difference between "ads" and "advertisements", for example, or "desks" and "furniture". Such a relationship between complexity and word length is clearly established:

> "One finding of cognitive science is that words have the most powerful effect on our minds when they are simple. The technical term is basic level. Basic-level words tend to be short. … Basic-level words are easily remembered; those messages will be best recalled that use basic-level language." (Lakoff & Wehling, 2012)

Finally, one can evaluate the mean sentence length (MSL). It has been observed that long sentences tend to render the speech more complex to understand. Table 2 depicts these three measurements individually for each president, and globally for both GPT versions. Moreover, in the last row, the average over the six presidents is shown by concatenating all their SOTU addresses. In this table, the largest values are presented in bold and the smallest in italics.

According to values shown in Table 2, GPT-3.5 presents the language with the highest complexity on the three measurements. On the other hand, Biden presents both the smallest mean of letters per word and the smallest MSL. Between the two GPT versions, one can observe that version 4.o clearly reduces the mean word size and the percentage of BW. Both values are still higher than the mean value over the six presidents (4.85 vs. 4.44, and 37.8% vs. 28.7%). The MSL of GPT-4.o corresponds clearly to possible presidential speech (19.95 vs. 19.45).

**Table 2.** Statistics on three language complexity measurements

|           | Mean word length | Big words | Mean sentence length |
|-----------|------------------|-----------|----------------------|
| GPT-3.5   | **5.07**         | **40.11%** | **21.56**           |
| GPT-4.o   | 4.85†            | 37.80%†   | 19.95†               |
| Reagan    | 4.49†‡           | 29.34%†‡  | 21.45‡               |
| Clinton   | 4.34†‡           | 26.79%†‡  | 21.33‡               |
| Bush      | 4.50†‡           | 30.13%†‡  | 20.07†               |
| Obama     | 4.31†‡           | *25.94*%†‡ | 19.72†              |
| Trump     | 4.55†‡           | 30.74%†‡  | 17.73†‡              |
| Biden     | *4.29*†‡         | 25.95%†‡  | *15.72*†‡            |
| Presidents| 4.44†‡           | 28.70%†‡  | 19.45†‡              |

To statistically determine whether a given mean could be viewed as different than that produced by GPT-3.5, a bilateral *t*-test (Conover, 1990) has been applied with the null hypothesis $H_0$ specifying that both population means are equal. For example, in Table 2 GPT-3.5 produces an average word length of 5.07 letters. Reagan pronounces on average 4.49 characters per word. This difference (5.07 – 4.49 = 0.58) must be viewed as statistically significant (significance level $\alpha$ = 1%), and this statistical significance is indicated by a single cross (†). Moreover, GPT-4.o presents a mean value of 4.85. This difference, compared with Reagan's mean, is also statistically significant (significance level $\alpha$ = 1%), and is denoted by a double cross (‡). With the BW values, the proportion test (Conover, 1990) has been applied instead of the *t*-test with the same significance level.

When comparing the two GPT versions, Table 2 shows that for the three measurements, GPT-4.o results in a lower language complexity, and the differences are always statistically significant compared to GPT-3.5. The newest version presents a reduced language complexity, closer but not similar to true presidents. As displayed in Table 2, the differences with GPT-3.5 are always statistically significant, as well as with the mean over all presidents. When comparing with GPT-4.0, the differences are usually always statistically significant.

When analysing a written style, the words can be divided into content and function terms with nouns, main verbs, adjectives and adverbs belonging to the first class. Function (or glue) words corresponding to pronouns, articles, prepositions, auxiliary verbs and conjunctions are more frequent and tend to reflect some stylistic characteristics. In particular, some stylistic and psychological traits of the author can be derived by analysing the relative frequencies of pronouns (Pennebaker, 2011; Kacewicz *et al*., 2014).

In this regard, the occurrence frequencies of personal and impersonal pronouns (e.g., it, that) (denoted *Ipron*[5]) are displayed in Table 3. The last row shows the percentage of pronouns when concatenating all presidential speeches and can be viewed as a mean usage for a president in power. As for the previous table, the largest values appear in bold and the smallest in italics. In addition, the proportion test has been applied with significant difference ($\alpha = 1\%$), denoted by † over GPT-3.5 or by ‡ over GPT-4.o.

---

[5] The term indicating a category is displayed in italics.

**Table 3.** Frequency of occurrence of pronouns

|  | *Self* | *We* | *You* | *She/he* | *They* | *Ipron* |
|---|---|---|---|---|---|---|
| GPT-3.5 | 1.05% | 6.93% | 0.56% | 0.00% | 0.77% | 3.59% |
| GPT-4.o | *0.68%*† | **8.30%**† | *0.46%*† | *0.00%* | *0.58%*† | *3.57%* |
| Reagan | 1.03%‡ | 4.25%†‡ | 0.50% | 0.23%†‡ | 0.84%‡ | 4.42%†‡ |
| Clinton | 1.55%†‡ | 4.45%†‡ | 0.77%†‡ | 0.33%†‡ | **1.31%**†‡ | 4.67%†‡ |
| Bush | 0.96%‡ | 4.11%†‡ | 0.64%†‡ | 0.29%†‡ | 1.18%†‡ | 3.69% |
| Obama | 1.32%†‡ | 4.28%†‡ | 0.55%‡ | 0.41%†‡ | 1.12%†‡ | **5.64%**†‡ |
| Trump | 1.16%‡ | 4.17%†‡ | 0.80%†‡ | **0.90%**†‡ | 0.93%†‡ | 3.73% |
| Biden | **1.98%**†‡ | *3.33%*†‡ | **1.37%**†‡ | 0.64%‡ | 1.26%†‡ | 4.65%†‡ |
| Presidents | 1.20%†‡ | 4.22%†‡ | 0.67%†‡ | 0.41%†‡ | 1.04%†‡ | 4.51%†‡ |

With the *Self* (I, me, mine, myself) category, GPT-4.o displays the smallest proportion of I-words while version 3.5 exposes a value close to that of some presidents (e.g., Reagan, Bush, or Trump). For a leader in an electoral campaign, a large proportion of *Self* corresponds to an efficient and successful communication strategy. After all, an election is the process of choosing between two candidates (e.g., US, Canada, France) (Labbé & Monière, 2008), (Savoy, 2018).

The use of we-words (we, us, our, ourselves) appear as a way to move from an individual point of view to a collective one, with a solidarity aspect. From a political communication point of view, this is a significant characteristic. The lemma 'we' is common to all political leaders in power. This pronoun has the advantage of being ambiguous; we are never sure who is behind the 'we'. Is it the president and his cabinet, the Congress, or more generally, the president and the people listening to the speech? In this last case, the speaker also wants to establish a relationship with the audience, usually to involve them in the proposed solution. As shown in Table 3, this pronoun is the most frequently employed by all presidents. Both versions of GPT overused it, and the proportion differences with all presidents are significant.

As shown in Table 3, GPT avoids using other personal pronouns. For GPT-4.o, those percentages are the lowest over all rows. One can explain these low rates by the difficulty of establishing the right reference between the referent and the pronoun. This is also true of the impersonal pronouns employed less frequently by the two GPT versions. Another finding is the absence of the third singular personal pronouns with GPT. More precisely, the word 'she' never appears under GPT's pen.

When analysing the differences between presidents, one can observe that Biden employs the lemma 'we' less frequently, but presents the highest intensity in the categories of *Self* and *You*. This choice denotes the willingness to establish a relationship between the speaker and the audience. These differences characterise Biden's voice as distinct from those of the other occupants of the White House.

When evaluating two or three personal pronouns, some psychological traits about the author can be perceived (Kacewicz *et al.*, 2014). People with higher status consistently use fewer first-person singular pronouns, and they use more first-person plural and second-person pronouns. The power language[6] is associated with attentional biases; higher status is linked with other-focus, whereas lower rank is linked with self-focus (Kacewicz *et al.*, 2014), (Pennebaker, 2011). According to this perspective, both GPT versions appear to adopt a high leader status with a high frequency of *We* and *You*, and a low percentage of *Self* (e.g., GPT4.o: 8.3% + 0.46% – 0.68% = 8.08%). Among presidents, the combined frequency of the categories *We* + *You* - *Self* indicates that Trump (3.81%) and Bush (3.79%) embrace a higher social status than the other presidents, with the lowest value associated with Biden (2.71%).

## 5 Psychological and Emotional Analysis

A psychological and emotional analysis of political speeches can be grounded on LIWC[7]. This text-based analysis system is built around several wordlists according to syntactical, emotional or psychological categories. The main hypothesis is to assume that the words serve as guides to the way the author thinks, acts, or feels (Jordan, 2022). In LIWC, categories may match grammatical categories such as personal pronouns, as well as broader ones (e.g., verbs), or more specific ones (verbs in the past tense, auxiliary verbs). On a semantics level, the LIWC defines positive emotions (*Posemo*) (e.g., happy, hope, peace), or negative ones (*Negemo*) (e.g., fear, blam*[8]). With these categories, the emotional aspect (optimism or pessimism) of a speaker can be evaluated. Presidents (or prime ministers) tend to voice positive words more frequently to appeal to the audience and to persuade the public. In particular, populist leaders more often employ emotional terms to incite strong sentiments in the population, usually to obtain a larger media coverage (Obradović *et al.*, 2020), (Hart, 2020), (Savoy & Wehling, 2022).

The category *Cogproc* contains terms related to self-reflection (e.g., think, refer*) and causal words (e.g., cause, understand). This measure corroborates with an active thinking and narrative tone (Tausczik & Pennebaker, 2010). Under *Achieve* (e.g., plan, win, lead*, etc.), one can evaluate the confidence of the author to resolve or to propose a solution to a problem in a successful way.

As a second approach, Hart *et al*. (2013) have developed the DICTION system, which groups different wordlists specifically created to analyse political messages. For example, in the *Familiarity* category (e.g., a, at, to, with, etc.), one can see words that occur in everyday expressions, and that correspond to terms which are easily understood (Ogden, 1968). Such an enumeration corresponds to a stopword list applied by search engines to ignore terms without a clear meaning (Dolamic & Savoy, 2010).  When opting for a high level of familiarity, the speaker wants to address his or her message to the entire population using a simple tone.  To reinforce this characteristic, the orator could present a lower mean number of letters per words and write short sentences (see Table 2).

---

[6] The power language is used by people higher in power and status (e.g., your boss).

[7] Linguistic Inquiry & Word Count (Tausczik & Pennebaker, 2010).

[8] When generating an entry in a wordlist, one can use the symbol '*' to denote any sequence of letters.

More specific to political text analysis, the category *Symbolism* contains terms related to the country (e.g., nation, America), ideology (e.g., democracy, freedom, peace), or generally political concepts and institutions (e.g., law, government).  Those expressions are related on an abstract level and are usually employed to express an ideal view of the situation.  Additionally, the *Politics* category (e.g., power, republican, majority, federal, etc.) contains concrete terms related to political institutions and parties in the US.

**Table 5.** Semantic categories over the US presidents and both GPT versions

|  | *Posemo* | *Negemo* | *Cogproc* | *Achieve* | *Familiarity* | *Symbolism* | *Politics* |
|---|---|---|---|---|---|---|---|
| GPT-3.5 | **7.34%** | 1.27% | 8.12% | **5.51%** | 20.06% | **5.24%** | 3.94% |
| GPT-4.o | 7.21%† | *0.99%†* | *7.23%†* | 4.36%† | *20.01%* | 5.37% | **5.40%†** |
| R. Reagan | 4.86%†‡ | 1.88%†‡ | 8.93%†‡ | 2.73%†‡ | **22.87%†‡** | 3.84%†‡ | 4.18%‡ |
| B. Clinton | 4.20%†‡ | 1.62%†‡ | 9.72%†‡ | 3.08%†‡ | 22.60%†‡ | 3.52%†‡ | 3.37%†‡ |
| W.G. Bush | 4.99%†‡ | **3.09%†‡** | 8.46%‡ | 2.91%†‡ | 21.93%†‡ | 4.10%†‡ | 4.19%†‡ |
| B. Obama | 3.66%†‡ | 1.73%†‡ | 10.31%†‡ | 2.86%†‡ | 22.17%†‡ | *3.16%†‡* | *3.10%†‡* |
| D. Trump | 4.29%†‡ | 2.34%†‡ | 7.63%† | 2.48%†‡ | 20.83%†‡ | 4.43%†‡ | 3.91%‡ |
| J. Biden | *3.33%†‡* | 1.74%†‡ | 9.11%†‡ | *2.09%†‡* | 21.32%†‡ | 3.50%†‡ | 3.31%†‡ |
| Presidents | 4.33%†‡ | 2.11%†‡ | 9.06%†‡ | 2.78%†‡ | 22.04%†‡ | 3.76%†‡ | 3.69%†‡ |

The percentages of each category achieved by the six presidents and the two GPT versions are reported in Table 5.  In the first two columns, both GPT versions employ more positive emotions and less negative ones compared to true presidents.  As one can see, the differences with the US leaders are always statistically significant.  Between presidents, Bush presents the highest percentages in both positive and negative feelings.  In particular, he obtains the highest negative score with terms related to the war in Iraq and terrorists.  One may be surprised to not see Trump with the highest percentage of negative terms.  This study is based on written speeches, certainly authored by ghostwriters and not the president himself.  With Trump, one can observe significant differences between his written messages and his spontaneous language (e.g., interviews, press conferences, tweets) (Savoy & Wehren, 2022).

With terms occurring in the *Cogproc* category, GPT-3.5 portrays a percentage similar to Bush.  Meanwhile, GPT-4.o, with the lowest value, is similar to Trump's percentage.  In this regard, Obama clearly shows the highest value.  For the categories *Achieve* and *Familiarity*, the differences are always significant with all of the presidents.  GPT more often uses terms in the *Achieve* class and less words appearing in the *Familiarity* one.  This finding confirms the presence of a complex formulation and longer words under GPT's pen.  Moreover, GPT opts for a tone which underlies accomplished or fulfilled tasks.

Both GPT versions employ more terms belonging to the *Symbolism* category, and the difference with the true presidents is always significant. Moreover, the difference in percentage between GPT-3.5 and GPT-4.o is not significant. When generating political texts, GPT favors words related to abstract ideas (e.g., freedom) and national references (e.g., America). Between presidents, Obama uses these terms less often.

When inspecting the percentages of terms appearing in the *Politics* category, the two GPT versions expose significant differences in their usage. The newest model displays the highest value, more frequently referencing concrete terms related to political institutions (e.g., Congress, state, president). The differences with the presidents are always significant.

Instead of focusing on a single percentage related to a given wordlist, the LIWC system proposes a combination of several categories to generate four composite measurements, namely emotional tone, confidence (or clout), analytical thinking, and authenticity. The resulting numbers are standardised scores based on some LIWC categories, and their values range from 1 to 100 (Pennebaker *et al.*, 2014; Jordan *et al.*, 2019). The computed values obtained with our corpus are depicted in Table 6, which shows the largest values in bold and the smallest in italics. Moreover, a bilateral *t*-test has been applied because the values correspond to the means over all of the SOTU addresses written by each president or GPT model.

The emotional *Tone* (Monzani *et al.*, 2021) combines both positive and negative dimensions (see also Table 5). Values larger than 50 indicate an overall positive tone, while numbers below this threshold are associated with an overall negative sentiment. As shown in Table 6, both GPT versions focus exclusively on a positive timbre. The differences with the true presidential allocutions are significant. In the latter case, one can observe both positive and negative terms. In majority, however, the positive ones dominate, in part because they must convince the citizens that they have the capacity to solve current problems, and that their actions are the most appropriate for the country. Moreover, they are pleased that they have the power. Finally, between presidents, Biden displays the lowest positive emotional tone (during his term, he was confronted with the COVID-19 pandemic and the war in Ukraine).

**Table 6.** Composite summary measurements (LIWC)

|  | *Tone* | *Clout* | *Analytical* | *Authenticity* |
|---|---|---|---|---|
| GPT-3.5 | **96.8** | 95.5 | 81.1 | 15.31 |
| GPT-4.o | **98.3**† | **97.3**† | 79.0† | *9.7*† |
| Reagan | 78.6†‡ | 85.3†‡ | 81.8‡ | 31.1†‡ |
| Clinton | 73.7†‡ | 89.3†‡ | 79.6†‡ | 32.6†‡ |
| Bush | 60.8†‡ | 89.3†‡ | **84.1**†‡ | 22.8†‡ |
| Obama | 62.0†‡ | 83.7†‡ | *71.7*†‡ | 37.1†‡ |
| Trump | 62.3†‡ | 89.7†‡ | 80.2†‡ | 30.0†‡ |
| Biden | *56.2*†‡ | *78.2*†‡ | 73.8†‡ | **40.0**†‡ |
| Presidents | 66.8†‡ | 86.6†‡ | 78.9† | 31.5†‡ |

The *Clout* (or confidence) category is used to determine the person's relative status in a social hierarchy. A leader must have a high status reflected by a higher usage of the pronouns 'we' and 'you' (see Table 3). On the contrary, a person of lower status tends to employ more I-words and impersonal pronouns (e.g., it, one) (Kacewitz *et al*., 2014; Pennebaker, 2011). People with a high social status present higher authoritative language and have a tone of higher certainty. As depicted in Table 6, both GPT versions expose a high value in this dimension. For both *Tone* and *Clout*, Biden shows the lowest value among US presidents.

The *Analytical* thinking measure has been shown to be associated with a greater academic level (Markowitz, 2023). This tone is grounded on a larger cognitive elaboration, leading to the impression of conveying more competence. An analytical language appears logical and formal, employs more articles and prepositions, and focuses more on noun phrases (Pennebaker *et al*., 2014; Jordan *et al*., 2022). Opting for a highly analytical tone, the speaker takes the risk of appearing too distant, impersonal, and lacking an emotional aspect. On the other hand, a more intuitive and personal person writes more often with pronouns, negations, auxiliary verbs, conjunctions and some adverbs (e.g., so, very) (Pennebaker *et al*., 2014). Among presidents, Bush presents the highest analytical thinking, while Obama expresses the lowest.

The *Authenticity* measurement (Pennebaker *et al*., 2014) is related to the way a leader is able to communicate in a spontaneous way (Markowitz *et al.*, 2023), a pitch usually viewed as an honest one. Adopting this characteristic, the language is more concrete and presents more self-references in a natural way. Leaders adopting this tone appear to be closer or more connected to people's interests (Hart, 2023). However, this attitude does not imply that the speaker tells the truth (Pennebaker, 2011). As displayed in Table 6, Biden presents the highest value, while both GPT versions depict the lowest values. All presidents expose a significantly higher score than both GPT versions.

From data depicted in Table 6, GPT has a highly positive emotional tone, adopts a high-power language, and lacks authenticity. Only in analytical thinking could GPT be viewed as a true president. Biden's image appears to be clearly distinct from that of other presidents, with a more negative tone that is both low in language power and analytical thinking, but that could be viewed as honest.

## 6 Intertextual Distance

To evaluate more globally the similarity between all presidents and both GPT versions, an intertextual distance between all pairs of texts can be computed (Labbé, 2007). The computation of this measure between Text A and Text B is defined according to the entire vocabulary. Equation 1 specifies this measure with $n_A$ indicating the length of Text A (in number of tokens), and $tf_{i,A}$ denoting the absolute frequency of the *i*th term (for $i = 1, 2, …, m$). The value *m* represents the vocabulary length. Usually, both texts do not have the same length, so we may assume that Text B is the longest. To reduce the longest text to the size of the smallest, each of the term frequencies (in our case $tf_{i,B}$) is multiplied by the ratio of the two text lengths, as indicated in the second part of Equation 1.

$$D(A, B) = \sum_{i=1}^{m} |tf_{i,A} - \widehat{tf_{i,B}}| \Big/ (2 \cdot n_A) \qquad \text{with } \widehat{tf_{i,B}} = tf_{i,B} \cdot n_A/n_B \qquad (1)$$

Having six presidents, and for each president the two GPT versions, one can count, in total, 18 texts. Directly displaying the 18 x 18 matrix containing these distances is of limited interest. Knowing that this matrix is symmetric and that the distance to itself is nil, we still have in total ((18 x 18) – 18) / 2 = 153 values. To achieve a better picture than a list of values or a dendrogram, such distance matrices can be represented by a tree-based visualisation *approximately* respecting the real distances between all nodes (Baayen, 2008; Paradis, 2011). We adopt this new representation, of which the result is displayed in Figure 1. Additionally, the string '35' has been added after each president's name to indicate speeches generated by GPT-3.5. A similar denomination has been applied for GPT-4.o.

**Figure 1.** Overall distance between presidents and GPT versions

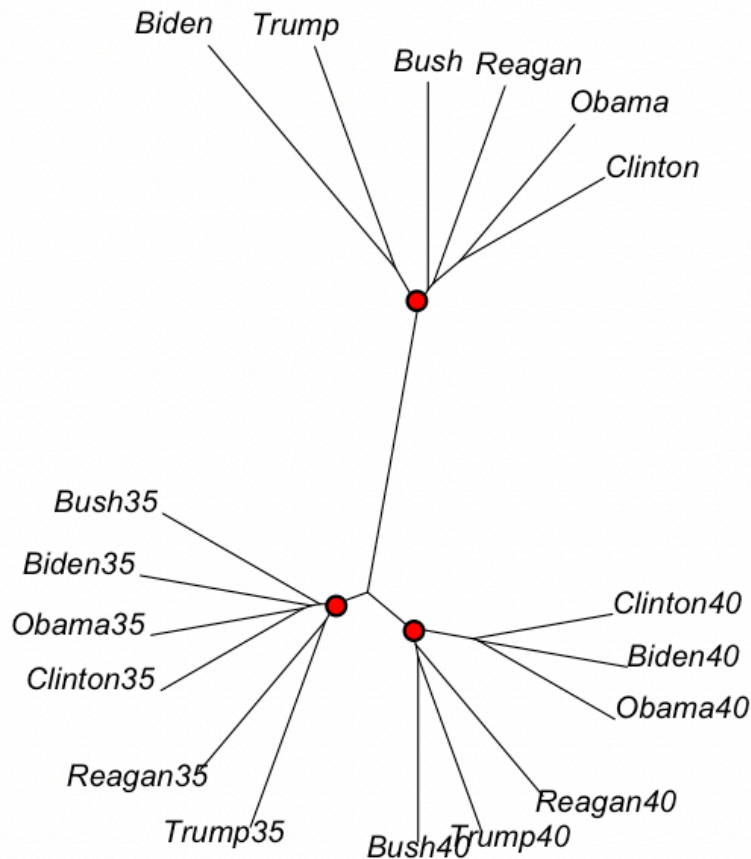

Overall, this figure illustrates the large difference between the true addresses (appearing on the top part) and the other GPT speeches (depicted in the bottom part). To obtain a better understanding of this picture, the starting point of each cluster is indicated by a red dot. The two

GPT versions clearly form two distinct subtrees, and the distance between them is smaller than with the set of true speeches.

With GPT-4.0, two subgroups can be defined: one with the Republican presidents (Bush 40, Trump 40, and Reagan 40), and a second with the Democrats (Clinton 40, Biden 40, and Obama 40). Moreover, the true presidents cluster displays a greater distance between each member than in the other two groups. Finally, one can observe that the last two US presidents (Trump and Biden) are displayed with some distance from the four others.

## 7  Conclusion

Some experiments performed in this study demonstrate that both GPT models can generate political speeches sharing some similarities with real *State of the Union* (SOTU) addresses. In addition, the newest version (GPT-4.o) exposes distinct characteristics compared to GPT-3.5. For example, the messages generated by GPT-4.o are significantly longer: on average, 645 tokens vs. 494 for GPT-3.5.

The two models share some common features, such as a higher language complexity compared to true presidents. In this regard, GPT generates longer words (the mean is 4.96 letters per word), with a higher percentage of big terms (on average, 39%), and longer sentences (20.76). Among presidents, Biden tends to present the lowest language complexity, with the shortest words and sentences.

When focusing on personal pronouns, both GPT versions opt for a large percentage of we-words (we, us, our) with few other pronouns (e.g., the third singular pronouns occur very rarely). Even if the increased frequency of we-words is a characteristic of political leaders in power, GPT employs them more often than true presidents. Between presidents, Biden presents a distinct figure with a relatively high number of I-words and second-person pronouns.

When inspecting emotional terms, both GPT models employ almost only positive terms (on average, 7.3%), leading to an optimist tone. True presidents also favour positive sentiments (on average 4.3%), along with some negative ones (2.1%). Among presidents, Bush writes with the highest number of emotional terms (on average, 4.99% are positive, 3.09% negative). This feature can be explained by the war in Iraq and against terrorists. Again, Biden uses the lowest percentage of positive terms (3.33%), and a low number of negative ones (1.74%).

When considering other categories, the two GPT versions opt for a larger percentage of *Achieve* (on average, 4.9%), *Symbolism* (5.3%), and *Politics* (4.7%) terms. This can be explained by the wish to anchor the speech in political parlance (e.g., nation, Congress, America) and to underline the results or actions already planned (e.g., win, plan). For the presidents, the average percentages are significantly lower (*Achieve:* 2.8%, *Symbolism:* 3.8%, *Politics*: 3.7%).

When considering other psychological measurements, both GPT models expose a clear language, belonging to a high-status person (*Clout*), but with a low value in authenticity. The resulting tone could appear authoritative and distant. Among presidents, Biden opts for a less optimistic and less confident tone that could also appears as being more honest.

Finally, by computing a global intertextual distance between each president and the corresponding messages generated by both GPT versions, one can observe three separate clusters: one for each GPT model, and one for the true presidents. Based on the language, the difference between machine-based speeches and real ones appears clearly, with GPT favouring a more complex language, opting for an optimistic feeling, and a more authoritative tone.

# Appendix

**Figure A.1.** Warning received from OpenAI when generating political speeches

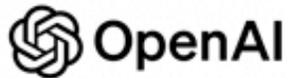

Hello,

OpenAI's Usage Policies restrict the use of our scaled services for political campaigning or lobbying. We've identified that your organization's use has resulted in requests that are not permitted under our policies. Your organization should immediately suspend use that violates those policies. If you have not remediated within three (3) calendar days, we may take additional action to suspend your access to our scaled services.

We will continually evaluate our approach as policymakers, members of civil society, and the public explore how our tools can empower people and solve complex problems. You can read more about the steps we are taking on elections here: How OpenAI is approaching 2024 worldwide elections.

Best,
The OpenAI team